\documentclass{article}

\usepackage[main, final]{template}
\usepackage[utf8]{inputenc} 
\usepackage[T1]{fontenc}    
\usepackage{hyperref}       
\usepackage{url}            
\usepackage{booktabs}       
\usepackage{amsfonts}       
\usepackage{nicefrac}       
\usepackage{microtype}      
\usepackage{xcolor}         
\usepackage{graphicx}
\usepackage{amsmath}
\usepackage{wrapfig}
\usepackage{subcaption}
\setcitestyle{authoryear,round,citesep={;},aysep={,},yysep={;}}

\title{The Representational Geometry of Number}

\author{%
  \textbf{Zhimin Hu}$^{1}$ \quad \textbf{Lanhao Niu}$^{2}$ \quad \textbf{Sashank Varma}$^{1}$ \\
  \vspace{1ex} \\
  $^{1}$ Georgia Tech \\
  $^{2}$ University of Edinburgh \\
}

\begin{document}

\maketitle

\begin{abstract}
  A central question in cognitive science is whether conceptual representations converge onto a shared manifold to support generalization, or diverge into orthogonal subspaces to minimize task interference. While prior work has discovered evidence for both, a mechanistic account of how these properties coexist and transform across tasks remains elusive. We propose that representational sharing lies not in the concepts themselves, but in the geometric relations between them. Using number concepts as a testbed and language models as high-dimensional computational substrates, we show that number representations preserve a stable relational structure across tasks. Task-specific representations are embedded in distinct subspaces, with low-level features like magnitude and parity encoded along separable linear directions. Crucially, we find that these subspaces are largely transformable into one another via linear mappings, indicating that representations share relational structure despite being located in distinct subspaces. Together, these results provide a mechanistic lens on how language models balance the shared structure of number representation with functional flexibility. It suggests that understanding arises when task-specific transformations are applied to a shared underlying relational structure of conceptual representations. \footnote{Code and data can be found at https://github.com/Paleuna/Number-Geometry}
\end{abstract}

\section{Introduction}
Representational geometry has emerged as a quantitative framework to investigate how the structure of representations supports intelligent behavior. A central tension in this area is the trade-off between generalization and task interference. On the one hand, \citet{huh2024position} suggests that as neural networks scale, they converge towards a shared ideal representation. On the other hand, empirical evidence suggests that systems often organize distinct tasks into task-specific subspaces to prevent interference \citep{yang2019task, libby2021rotational}. \citet{bernardi2020geometry} proposes that neural representations resolve the trade-off by expanding representations into a high-dimensional space to make complex information linearly accessible, while ensuring that the underlying abstract categories are compressed into a low-dimensional structure.

Despite this progress, two significant gaps remain. First, existing research lacks a dynamic perspective; representations are often studied in static contexts or within a narrow class of tasks. Second, we lack a mechanistic account of how task subspaces are transformed into each other and how they relate to shared information (if this exists).

In this paper, we argue that the shared nature of neural representations lies not in the concepts themselves, but in the geometric relations between them. We hypothesize that task-specific representations can be understood as transformations applied to this common structure, and that they are projected into distinct subspaces to avoid task interference.

To investigate this proposal, we utilize number concepts as a testbed and large language models (LLMs) as computational instances. Numbers are uniquely suited as they are well-defined for various tasks and their representations, such as the Mental Number Line (MNL), are supported by a rich psychophysical literature \citep{dehaene2001precis}. LLMs, as computational instances, provide rich high-dimensional representational spaces and show promising alignment with human cognition and its development \citep{shah2024development}.

We analyze a diverse set of LLMs spanning different architectures, training objectives, and scales--from BERT \citep{devlin2019bert} and GPT-2 \citep{radford2019language} to Qwen2.5-7B and its math-specialized variant \citep{qwen25technicalreport, yang2024qwen2}. We extract contextual embeddings of numbers across numerical tasks as task-specific conceptual representations. To examine their structure, we apply a sequence of representational analysis methods, including Procrustes Analysis \citep{gower1975generalized}, Subspace Overlap, and Canonical Correlation Analysis \citep{morcos2018insights}.

Our investigation finds that across tasks and models, the relative geometric relationships between numbers remain largely consistent, and such relations largely match the known structure of the MNL. To minimize interference, models utilize distinct subspaces to represent different tasks. Low-level features, such as parity, are encoded in nearly orthogonal directions to numerical magnitude. While task subspaces occupy different representational directions, they show similar relational structure. Many task-specific representations can be mapped to each other via linear transformations.
Together, these results reveal the representational geometry of number.

\section{Related Work}

\subsection{Representational Geometry}
Early theories of representation emphasized decomposing concepts into independent parts. \citet{higgins2018towards} formalized disentanglement via group theory, proposing that ideal representations mirror the world’s symmetries (e.g., object identity remains invariant under rotation). Similarly, \citet{park2024linear} argued that concepts in LLMs correspond to linear directions in latent space, with semantic opposites (e.g., man vs. woman) lying in orthogonal directions.

These perspectives have influenced two major lines of research. The first explores how systems mitigate task inference. \citet{yang2019task} found that neural networks trained on multiple cognitive tasks spontaneously cluster representations into functionally specialized regions. \citet{libby2021rotational} showed that neural circuits reformat sensory inputs via rotational dynamics into orthogonal memory subspaces over time. The second line focuses on generalization. \citet{bernardi2020geometry} argues that neural representations adopt geometries that support high shattering dimensionality to enable task flexibility through linear decodability, while constraining abstract categories to low-dimensional structures for generalization. Most recently, \citet{huh2024position} posits that as models scale, they tend to converge on a shared, ideal representation of reality, invariant to architecture or modality. 

These works characterize what representational geometries look like under various constraints. What has been lacking is a mechanistic account of how they dynamically transform across tasks.

\begin{figure*}
  \centering
  \includegraphics[width=1\textwidth]{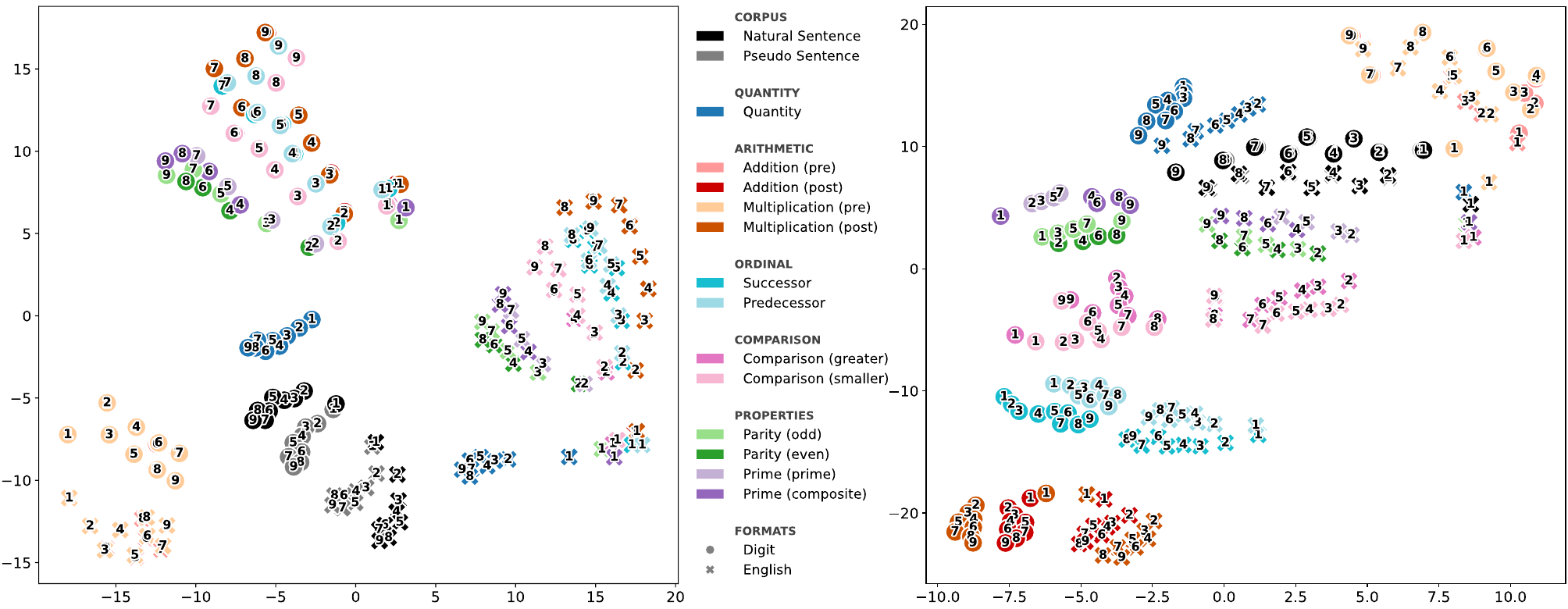} 
  \caption{t-SNE projections of number representations extracted from BERT (left) and Qwen2.5-Math (right) across numerical tasks. Each point corresponds to a representation of a number (1–9). Colors indicate task categories, while marker shapes represent formats (digit vs. English word).}
  \label{fig:tsne}
\end{figure*}

\subsection{Numerical Cognition}
Human numerical cognition is understood as comprising two systems: the Approximate Number System, an evolutionarily ancient system to estimate magnitudes, and the Symbolic System, which enables exact calculation and rule-based manipulation through culturally acquired symbols \citep{cantlon2012math, ansari2016number}. The highly influential Triple-Code Model proposes that numerical information is represented in three distinct formats -- Arabic (i.e., digits), verbal (i.e., words), and analog magnitude (i.e., MNL) -- each associated with a different neural substrate \citep{dehaene2001precis}. These systems interact dynamically during numerical tasks, mapping between codes as needed.

In particular, the analog magnitude representation is viewed as a central code for supporting magnitude estimation and comparison. It gives rise to three behavioral effects that collectively imply a core representational structure for number: a logarithmically compressed MNL. The Distance Effect is that the time to compare which of two numbers is greater increases with numerical distance between them (e.g., 1 vs. 9 is compared faster than 1 vs. 2) \citep{moyer1967time}. The Size Effect is that, for a fixed distance, smaller numbers are compared more quickly than larger ones (e.g., 1 vs. 2 is faster than 8 vs. 9) \citep{parkman1971temporal}. Finally, the Ratio Effect is that comparison time decreases as the ratio between the numbers increases \citep{halberda2008individual}. \citet{shah2023numeric} has shown that LLMs capture these effects in their non-contextual embeddings. However, they do not address how number representations transform across different contexts.

\section{Methods}

\subsection{Task Design and Material} 
To examine the representational geometry of numbers across tasks, we chose a set of numerical tasks from the mathematical cognition literature applied to the numbers 1–9. Numbers were presented in two formats: Digit (e.g., “3”) and English words (e.g., “three”). For each task, we crafted five sentences that elicited a specific cognitive operation. These sentences were fed into LLMs, and we extracted internal activation vectors (embeddings) corresponding to the number token (shown in \textbf{bold} below). The tasks include the following:

\begin{itemize}
\item \textbf{Quantity}: Represents the cardinality of sets--the mapping between numerical symbols and discrete set sizes.\\
\textit{Quantity: e.g., ``I have a total of \textbf{3} apples.''}

\item \textbf{Comparison}: Relates to magnitude ordering, evaluating the relative size of numbers on a scale. It includes both the case when a number is \textit{greater} than or \textit{smaller} than the other. \\
\textit{Comparison (greater): e.g., ''A number bigger than two is \textbf{three}.''}\\
\textit{Comparison (smaller): e.g., ''A number lower than two is \textbf{one}.''}

\item \textbf{Arithmetic}: Relates to algebraic operations involving both addition and multiplication, in two variants: \textit{pre-equal} (target as operand) and \textit{post-equal} (target as result).\\
\textit{Multiplication (pre-equal): e.g., ``The product of 1 and \textbf{3} is 3.''}\\
\textit{Addition (post-equal): e.g., ``Zero added to three gives \textbf{three}.''}

\item \textbf{Properties}: Relates to number properties, identifying discrete mathematical categories such as parity and primality.\\
\textit{Primality: e.g., ``Example of prime value is \textbf{5}.''}\\
\textit{Parity: e.g., ``The set of odd numbers includes \textbf{three}''}

\item \textbf{Ordinal}: Relates to numerical sequence, identifying the position of a number relative to its immediate neighbors ($n \pm 1$).\\
\textit{Successor: e.g., ``The number after two is \textbf{three}.''}\\
\textit{Predecessor: e.g., ``The number before four is \textbf{three}.''}
\end{itemize}

To evaluate number representations in naturalistic settings, we constructed two sentence corpora. \textbf{Pseudo Sentence} contains 900 seven-word segments generated by chunking Wikipedia text, with the numbers randomly inserted 100 times each (e.g.,\textit{``When \textbf{one} she was put off service and...''}); \textbf{Real Sentence} comprises 450 naturally occurring Wikipedia sentences containing the numbers, similarly chunked (e.g.\textit{``After a \textbf{2} year long engagement, Hilda...''})

\subsection{Measures, Metrics, and Analysis Strategies}
To quantitatively assess the geometric structure of numerical representations, we first examine three effects that signal recruitment of magnitude representations (i.e., an MNL). Following \citet{shah2023numeric}, we use cosine similarity as a proxy for reaction time. For each number pair $(i,j)$, the distance effect is quantified by fitting a linear model to similarity vs. $|i-j|$; the size effect by regressing normalized similarity on $\min(i,j)$; and the ratio effect by fitting a negative exponential function to similarity as a function of $x = max(i,j)/min(i,j)$. 

We also apply Procrustes Analysis \citep{gower1975generalized} to estimate the degree to which the geometric relations between numbers are shared. Given two embeddings $X$ and $Y$, it seeks a rigid transformation (translation, rotation, and uniform scaling) to minimize the: 
$$M^2 = \min_{s, \mathbf{R}, \mathbf{t}} \| s\mathbf{X}\mathbf{R} + \mathbf{1}\mathbf{t}^\top - \mathbf{Y} \|^2_F$$
In this formulation, $s$ is a scalar scaling factor, $\mathbf{R} \in \mathbb{R}^{n \times n}$ is an orthogonal rotation matrix satisfying $\mathbf{R}^\top\mathbf{R} = \mathbf{I}$, and $\mathbf{t}$ is a translation vector. The resulting $M^2$, referred to as the Procrustes disparity, quantifies the residual geometric distortion. A smaller $M^2$ indicates a higher degree of structural isomorphism between the two latent spaces.

To assess whether one task spans the same subspace as others, we apply asymmetric Subspace Overlap Analysis. 
This evaluates how well the top-$k$ principal components  \citep{abdi2010principal} of task A explain variance in task B. Let $\mathbf{V}_A \in \mathbb{R}^{d \times k}$ be the matrix whose columns are the top-$k$ orthonormal principal components of task $A$, and $\mathbf{X}_B$ represent the centered representation matrix of task $B$. The overlap is formalized as:$$\text{Overlap}(A \leftarrow B) = \frac{\sum_{i=1}^k \text{Var}(\mathbf{X}_B \mathbf{v}_{A,i})}{\text{Var}(\mathbf{X}_B)}$$Here, $\mathbf{v}_{A,i}$ denotes the $i$-th principal component of $A$. The numerator represents the sum of the variances of task $B$ when projected onto each of $A$'s top-$k$ components, while the denominator $\text{Var}(\mathbf{X}_B)$ is the total variance of $B$.

To quantify the functional equivalence between two representations under linear transformations, we employed Singular Vector Canonical Correlation Analysis (SVCCA; \citealp{raghu2017svcca}). This first applies Principal Component Analysis (PCA) \citep{abdi2010principal} to both tasks to retain their top $n$ components, $\mathbf{PC}_{A}$ and $\mathbf{PC}_{B}$. Canonical Correlation Analysis (CCA) \citep{morcos2018insights} then identifies pairs of projection vectors $(\mathbf{w}_{A,i}, \mathbf{w}_{B,i})$ that maximize the Pearson correlation:$$\rho_i = \max_{\mathbf{w}_{A,i}, \mathbf{w}_{B,i}} \frac{\text{Cov}(\mathbf{w}_{A,i}^\top \mathbf{PC}_{A}, \mathbf{w}_{B,i}^\top \mathbf{PC}_{B})}{\sqrt{\text{Var}(\mathbf{w}_{A,i}^\top \mathbf{PC}_{A}) \cdot \text{Var}(\mathbf{w}_{B,i}^\top \mathbf{PC}_{B})}}$$Each $\rho_i$ is a canonical correlation coefficient, representing the maximum correlation between linear combinations of the two sets of components that remain orthogonal to all previously identified $i-1$ pairs. This yields a sequence $\rho_1, \rho_2, \dots, \rho_n$, where the mean value $\bar{\rho} = \frac{1}{n}\sum_{i=1}^n \rho_i$ reflects the overall functional similarity between the task representations.


\subsection{Models}
To examine how conceptual representations differ across architectures, training objectives, and domains, we selected four representative LLMs in their base (pre-trained only) forms
We included BERT \citep{devlin2019bert} and GPT-2 \citep{radford2019language} to isolate the effect of training objectives. BERT is trained to predict masked tokens utilizing bidirectional context (left and right). In contrast, GPT-2 learns representations by predicting only the next token. Qwen2.5 \citep{qwen25technicalreport}, a modern high-capacity model, offers insight into number concepts in higher-dimensional space. Qwen2.5-Math \citep{yang2024qwen2}, built on Qwen2.5, is further pre-trained on mathematical corpora, allowing us to assess how domain-specific training reshapes numerical representational geometry. 

\section{Results}
We primarily use representations from the 75\% depth of the layers for all experiments, as middle-to-late layers are generally found to exhibit stable brain alignment and semantic structure \citep{caucheteux2022brains}. We present results for BERT and Qwen2.5-Math, as they are representative of the broader trends observed across BERT, GPT-2, and Qwen2.5. Embeddings are scaled by L2 normalization. For most analyses (unless otherwise noted), each number representation we computed is the mean of the embeddings for each number (in Digit format) across tasks.

\begin{figure*}
  \centering
  \includegraphics[width=1\textwidth]{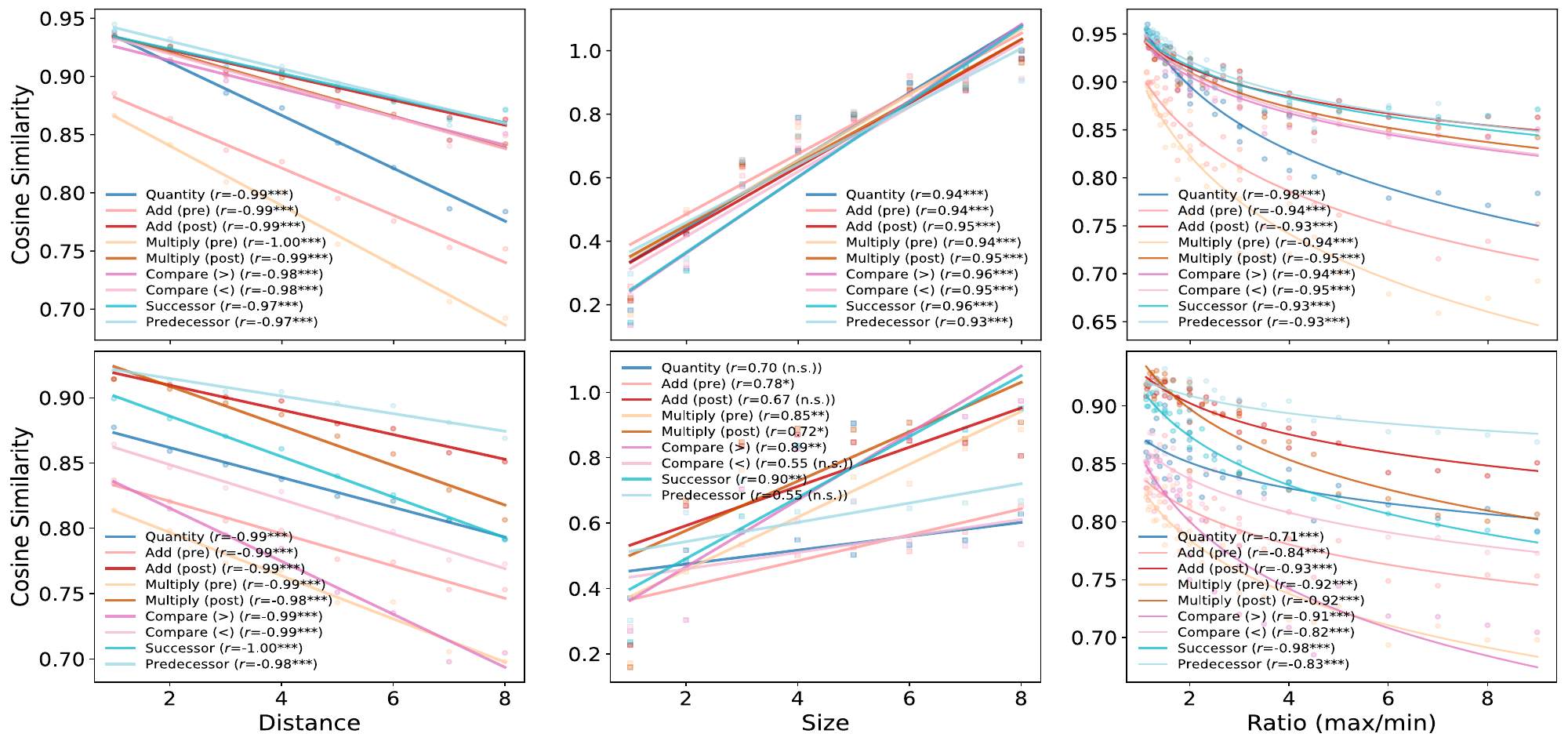} 
  \caption{Three effects fitted across models. Cosine similarity between number embeddings (Digit) is plotted against numerical distance (left), size (middle), and ratio (right) for each task. Top row: BERT; bottom row: Qwen2.5-Math. Each line represents a linear (left/middle) or logarithmic (right) fit for a given task (color-coded), with absolute Pearson’s r and significance level indicated in the legend ($p<0.05:$ *, $p<0.01:$ **, $p<0.001:$ ***).}
  \label{fig:three_effect}
\end{figure*}

\subsection{Shared Geometric Structure}
We first examine whether numerical concepts are embedded in a consistent geometric structure across tasks, number formats, and models.

To visualize the global organization of embeddings among tasks, we projected them into a 2D space using t-SNE \citep{hinton2002stochastic}. As shown in Figure~\ref{fig:tsne}, both models exhibit a highly organized layout where numbers 1–9 maintain a consistent sequential order within their respective task clusters. In BERT, the Comparison, Ordinal, and Arithmetic tasks show significant overlap in their geometric layout. Furthermore, the two number formats occupy distinct regions of the latent space, suggesting that semantic identity does not fully converge--a finding contrary to the Platonic hypothesis \citep{huh2024position}. Interestingly, their structures appear to reflect a diagonal axis, implying that while the model distinguishes number formats, it applies a shared relational structure to the underlying numerical concepts. In contrast, Qwen2.5-Math demonstrates a more fine-grained task segregation. For instance, Addition and multiplication are separated in the post-equal condition and form seemingly symmetric clusters, and tasks/properties such as parity and primality are more linearly separable. 

Meanwhile, we find that the distance, size, and ratio effects are largely preserved across tasks.\footnote{Note that we excluded the parity and primality tasks from this analysis as they exhibit distinct structures; see the next section.} As shown in Figure~\ref{fig:three_effect}, both models exhibit a near-perfect fit for the distance and ratio effects, indicating that the relative positioning of numerical concepts remains consistent and is sensitive to proportional differences between values. However, for the size effect, while BERT shows a stable fit across all tasks, Qwen2.5-Math exhibits greater variance. In particular, both the “greater than” comparison and the successor task preserve the size effect, whereas their counterparts do not. A similar pattern appears in the “pre-equal” vs. “post-equal” conditions of arithmetic. These results suggest that Qwen2.5-Math develops more fine-grained task-specific distinctions and prioritizes logical structure over general magnitude encoding. Notably, the size effect exhibits similar patterns across task groups. For example, the “greater than” comparison task shows trends that parallel those of the successor, indicating that task-specific logic is embedded similarly. 

To assess whether the geometric structure of number representations is a shared property across models, we applied Procrustes Analysis \citep{gower1975generalized}. Across all models, we observed a remarkably low average disparity between task-specific subspaces ($mean \approx 0.010$). To establish the significance of this alignment, we generated a permutation baseline by randomly shuffling numerical identities, which yielded significantly higher disparities (ranging from $0.077$ to $0.273$). This substantial gap confirms that the observed geometric alignment is highly non-random, suggesting that a consistent relational scaffold for numbers emerges across fundamentally different model families.

These findings reveal a structured spatial organization of representations and clear task-wise clustering. Despite some model-specific variation, all of the tasks shown in Figure~\ref{fig:three_effect} largely exhibit the three effects, supporting the view that they encode relative numerical relationships in a human-like manner. Further, Procrustes Analysis suggests that it is this relational structure--rather than specific coordinates--that constitutes the shared property across tasks.

\begin{figure*}
  \centering
  \includegraphics[width=1\textwidth]{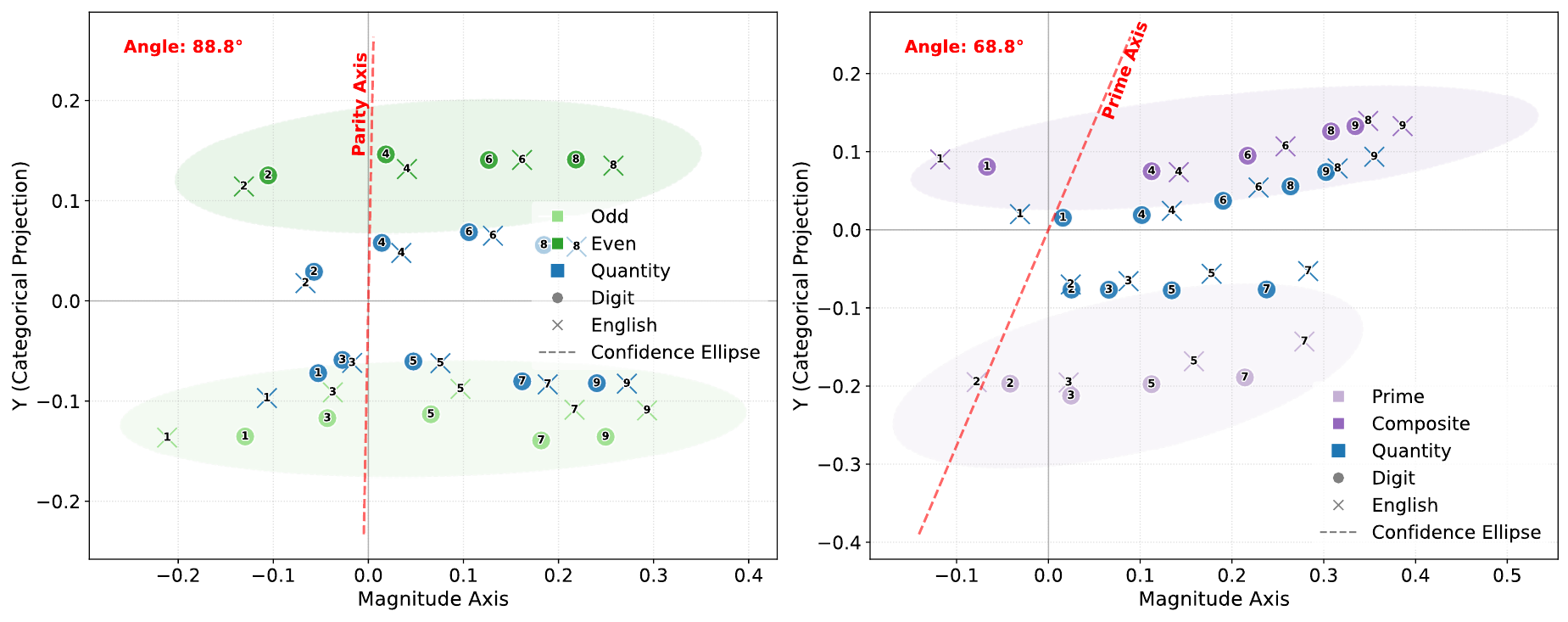} 
  \caption{Disentanglement of magnitude and parity/primality in Qwen2.5-Math. The x-axis represents the magnitude axis, and the red dashed line represents the categorical axis (odd vs. even left, prime vs. composite right). The y-axis shows orthogonal projection for visualization only. Circles and crosses denote formats; shaded ellipses mark 95\% confidence regions of raw embeddings.}
  \label{fig:parity}
\end{figure*}

\subsection{Orthogonality and Linear Equivariance}
After confirming the shared relational structure, it is unclear how the representational geometry supports task flexibility. As Qwen2.5-math demonstrated fine-grained task groups, we further examine how it maintains functional properties of numbers without interference. 

To explore the encoding of low-level features like magnitude and parity, we visualize the subspace of the parity and primality task by defining a 2D projection on the raw embeddings (instead of the mean). In this projection, one axis captures numerical magnitude (via linear regression), and the other reflects categorical distinctions--parity (odd vs. even) or primality (prime vs. composite)--derived from the mean difference between group centroids. As shown in Figure~\ref{fig:parity}, the parity axis is nearly orthogonal to the magnitude axis (88.8°), while the primality axis shows a noticeable tilt (68.8°). For parity, numbers cluster symmetrically into odd groups (below) and even groups (above) regardless of input format (digit or word), indicating that parity is encoded as a directionally independent feature. In contrast, the representation of primality shows partial coupling with magnitude, consistent with the fact that composite numbers become denser as their values increase. It reveals that the model encodes numerical properties along distinct linear directions.

\begin{figure*}
  \centering
  \includegraphics[width=1\textwidth]{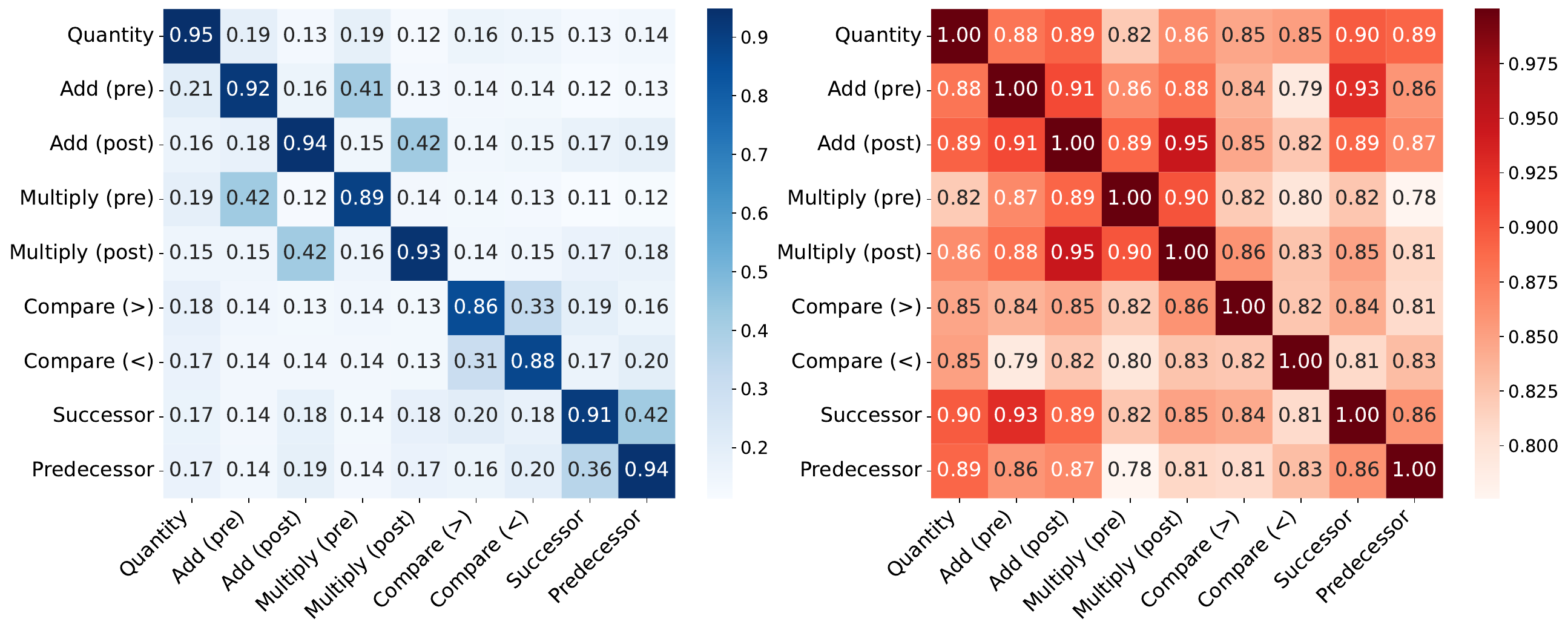} 
  \caption{Task-specific representation (Digit) alignment in Qwen2.5-Math. The left panel shows an asymmetric Subspace Overlap Ratio, while the right panel shows symmetric SVCCA similarity.}
  \label{fig:cca}
\end{figure*}

Moving from low-level features to functional relationships, we further analyzed the relationships between task-specific subspaces using Subspace Overlap Analysis and SVCCA. For both methods, we used raw embeddings and set the number of principal components to the averaged number of components needed to explain 95\% of the variance across tasks. As shown in Figure~\ref{fig:cca}, most task pairs exhibit minimal subspace overlap (typically < 0.2), with notable exceptions only among functionally related tasks such as Successor and Predecessor (0.36–0.42). This indicates that the model encodes tasks in highly decoupled subspaces. Despite this low physical overlap, SVCCA reveals consistently high canonical correlations (typically 0.80–0.90) across task pairs, suggesting that although subspaces are directionally separated to reduce interference, they retain a shared representational structure. Notably, the representations for parity and primality show lower SVCCA alignment compared to other tasks. This reduction in correlation implies that as task complexity increases, subspaces become more structurally distinct and cannot be fully aligned via simple linear transformations--pointing to the emergence of additional, possibly non-linear, structure. It confirms that representations occupy different subspaces to minimize inference and maintain the efficiency of task-switching via largely linear--and occasionally non-linear--transformations.

The disentanglement observed in parity and primality tasks suggests that the model encodes low-level numerical features along linearly separable directions. Yet, despite orientation changes, the overall structure of numerical relationships remains preserved, as indicated by moderate alignment between the two and quantity tasks in SVCCA. Together, the low subspace overlap and high (but variable) SVCCA scores support our central claim: numerical representations in LLMs are not fixed coordinates but stable, transformable geometries. While the model uses distinct subspaces to minimize task interference, it balances the shared structure and efficient task-switching primarily through linear transformation equivariance.

\begin{wrapfigure}{l}{0.5\textwidth} 
  \centering
  \includegraphics[width=0.49\textwidth]{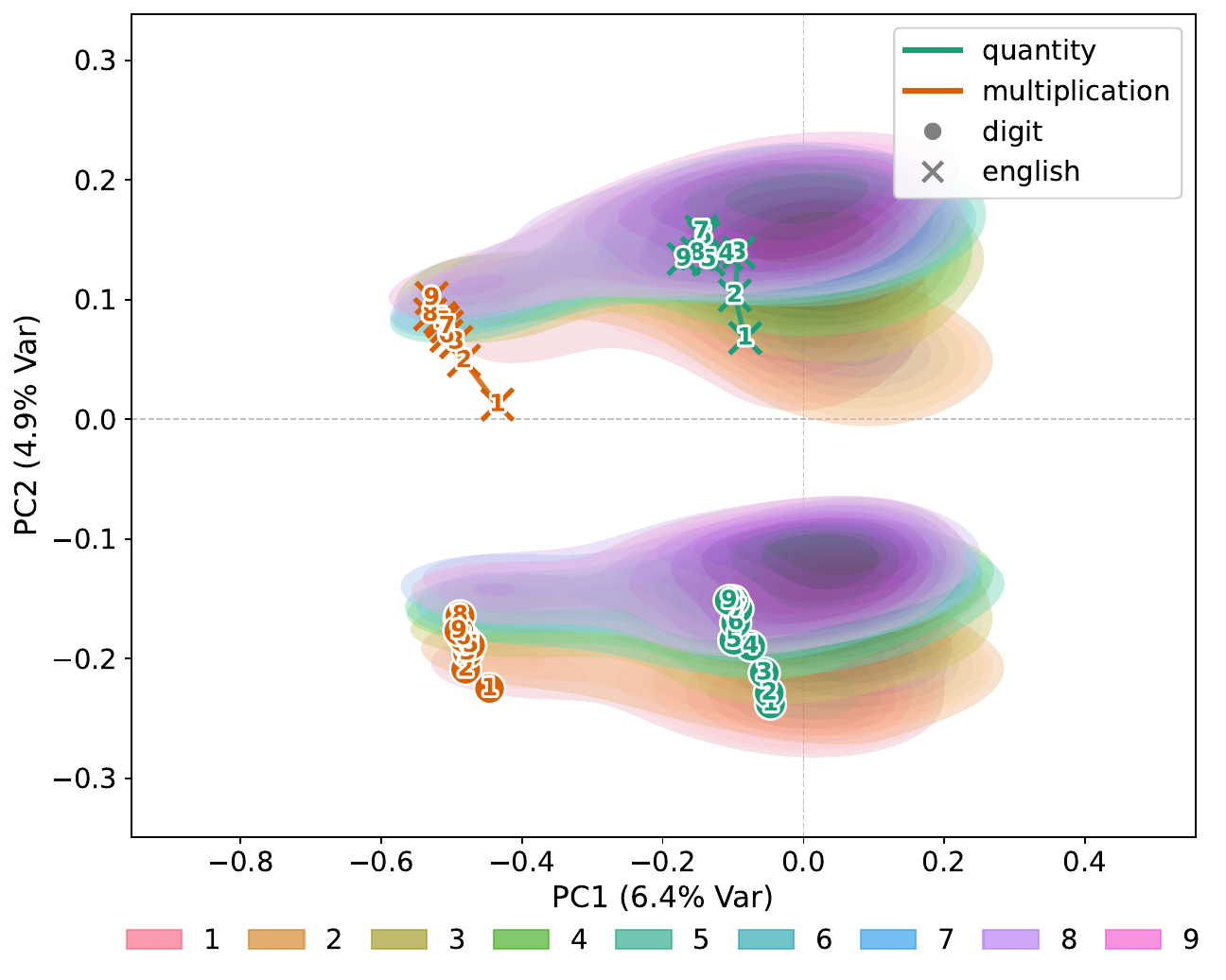}
  \caption{Distribution of number concepts in Qwen2.5-Math. Raw embeddings are projected using PCA and visualized using KDE (top 80\% density).}
  \label{fig:kde}
\end{wrapfigure}

\subsection{Distribution of Number Concepts}

To close, we provide a global distribution of Qwen2.5-Math's numerical concept space under more natural semantics. We project raw embeddings (instead of the mean) of numbers from both the corpus and handcrafted task sentences using PCA and visualize their concentration with kernel density estimation \citep{rosenblatt1956remarks}. This is a nonparametric method that yields a smooth estimate of where samples cluster in the projected space. Despite the limited variance captured in 2D, the projected distributions exhibit a stable ordinal organization across numbers, consistent with the shared relational structure discussed; see Figure~\ref{fig:kde}. Crucially, highlighted trajectories (e.g., Quantity vs. Multiplication) occupy distinct regions within this distribution, echoing our subspace-level results.

Furthermore, the near-parallel layouts for Digit and English inputs further indicate that the same relational structure is preserved despite being processed through separate representational channels. This separation is achieved without altering the distributional properties of the representations. We quantified this using representation sparseness \citep{rolls1995sparseness}, which measures how neural activity is distributed across the available dimensions. We found that sparseness remains largely similar across formats and numbers (Digit: $0.137 \pm 0.007$; English: $0.156 \pm 0.012$). It suggests that the model maintains a consistent level of feature concentration, which echoes the consistent topological density shown in Figure~\ref{fig:kde}. Together with the previous results, it implies that the model does not merely reuse the shared relational structure but preserves a consistent coding resolution.

\section{Discussion}

In this study, we show that number concepts are represented as structured geometric relations. Number features such as magnitude and parity are reliably encoded as linear directions in the representational space. Each task spans a distinct subspace, yet they are largely related through linear transformations, preserving similar underlying relational structure. \footnote{These results are not confined to a specific layer of models. We observe that both middle and late layers consistently preserve the geometric organization.} This provides support for \citet{gardenfors2004conceptual}’s conceptual space theory, yet goes beyond the idea of concepts as regions in a structured metric space as relations in vector space \citep{piantadosi2024concepts}.

Our findings organize previous theories of representational geometry--including the Platonic representation hypothesis \citep{huh2024position}, the linear representation hypothesis \citep{park2024linear}, and the task-disentangled subspace perspective \citep{yang2019task}. While we differ in emphasis, our account unifies them under a common relational perspective and has the potential to extend beyond number concepts. These findings also bear on a foundational question in neural systems and representation: if the same behavior or structure can arise from multiple implementations \citep{hu2025function}, what constrains representational convergence? One answer is the interplay between system capacity and environmental complexity. In small systems, representational redundancy often impedes the emergence of a stable geometry. However, as systems face increasingly complex learning environments--much like the human brain--they are pressured to converge to a more structured coding scheme. This suggests that knowledge representation is not necessarily innate \citep{spelke2007core}, but is a learned structural adaptation to the demands of the data.

Unlike the human brain, LLMs are not constrained by biological wiring or evolutionary contingencies. Rather, they are an unconstrained substrate on which representations can scale more freely. As such, they offer a powerful tool for probing what kinds of representational structures are necessary for thought, and also how and why such structures may succeed or fail to emerge in humans. For example, numerical cognition in humans is implemented over anatomically distributed systems \citep{dehaene2001precis}. If one system lacks sufficient scale or connectivity, it may fail to converge onto the expected geometric abstraction. This may help explain why individuals with dyscalculia--whose intraparietal sulcus (IPS) often shows reduced gray matter volume or atypical connectivity \citep{butterworth2011dyscalculia}--struggle with higher-order number representations.

\bibliographystyle{unsrtnat}
\bibliography{bibliography}

\end{document}